# Dynamical Systems Trees

**Andrew Howard**
Department of Computer Science
Columbia University
New York, NY 10027
*ahoward@cs.columbia.edu*

**Tony Jebara**
Department of Computer Science
Columbia University
New York, NY 10027
*jebara@cs.columbia.edu*

## Abstract

We propose dynamical systems trees (DSTs) as a flexible class of models for describing multiple processes that interact via a hierarchy of aggregating parent chains. DSTs extend Kalman filters, hidden Markov models and nonlinear dynamical systems to an interactive group scenario. Various individual processes interact as communities and sub-communities in a tree structure that is unrolled in time. To accommodate nonlinear temporal activity, each individual leaf process is modeled as a dynamical system containing discrete and/or continuous hidden states with discrete and/or Gaussian emissions. Subsequent higher level parent processes act like hidden Markov models and mediate the interaction between leaf processes or between other parent processes in the hierarchy. Aggregator chains are parents of child processes that they combine and mediate, yielding a compact overall parameterization. We provide tractable inference and learning algorithms for arbitrary DST topologies via an efficient structured mean-field algorithm. The diverse applicability of DSTs is demonstrated by experiments on gene expression data and by modeling group behavior in the setting of an American football game.

## 1 INTRODUCTION

Dynamical Bayesian networks are popular instantiations of graphical models that have shown promise in many applied settings such as computational biology, speech, and vision. Recently, graphical models and approximate inference methods have extended traditional dynamical systems, improving upon classical linear Kalman filters and hidden Markov models (HMMs) and exploring couplings and interactions between multiple hidden Markov chains. Such extensions include factorial HMMs [2] which indirectly link multiple Markov chains through a common output emission stream (Figure 1(a)). Meanwhile, coupled HMMs [8] directly link hidden states of multiple interacting processes that have Markovian temporal dynamics which generate different output emission streams (Figure 1(b)). Other extensions involve linking discrete and continuous Markov chains through so-called switching dynamical systems (SLDSs) that combine Kalman filters and HMMs [9, 1, 7, 5] to obtain nonlinear continuous dynamics (Figure 2). All the above systems basically link (directly or indirectly) hidden Markov chains together so they can influence each other in time. But, unlike simpler models, these variants involve hard inference and may require sampling [7] or structured mean field approximations [11].

In this article we propose a novel variant of dynamical systems for characterizing interacting processes that form groups and sub-groups. For instance, consider a football game where players are each modeled as a switching dynamical system. Players could interact with other members of their team through a hidden parent team state which has its own Markovian dynamics. The other team has its own Markovian team state which couples its players. Finally, an overall game state is a parent of and couples the two team states mediating their interaction. We call this model a dynamical systems tree (DST) since it permits the interacting processes to couple to each other by being mediated through an arbitrary tree hierarchy of aggregating hidden processes. The DST's arbitrary hierarchical tree structure has high level aggregating hidden states coupling groups of hidden Markov states that are parents of sub-groups of leaf dynamical systems (i.e. SLDSs or HMMs). Since higher level Markov chains are parents of lower level chains, a recursive structured mean field algorithm for inference is easy to derive for arbitrary tree structures and group/sub-group arrangements for the interact-ing processes. Thus, we are free to consider various ways that the individual leaf dynamical systems interact in a group scenario. This article describes and motivates the generative model for DSTs. Parameter estimation for DSTs is then derived via Expectation-Maximization and a structured mean field inference algorithm which can be applied recursively on any DST topology. Our structured mean field algorithm is more computationally efficient than those proposed by related models [9, 2]. We then provide and discuss promising experimental results with DSTs on gene expression data and on trajectories of players from real football data.

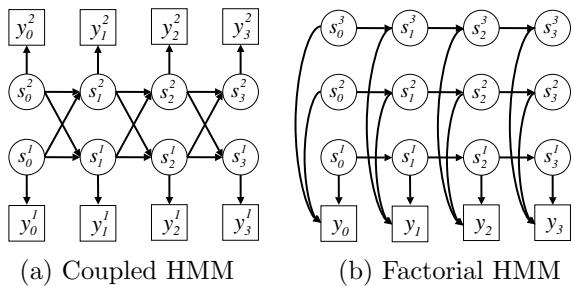

(a) Coupled HMM　　(b) Factorial HMM

Figure 1: Interacting Dynamical Systems.

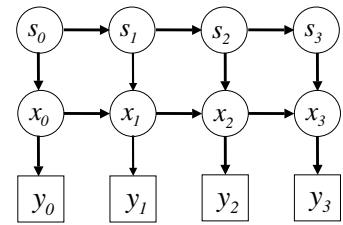

Figure 2: A Switching Linear Dynamical System.

## 2　DYNAMICAL SYSTEMS TREES

Instead of modeling interaction by direct coupling (as in coupled HMMs) or through shared outputs (as in factorial HMMs), we propose that processes interact through parent hidden Markov chains that act as mediators or aggregators of the sub-processes. These parent chains have their own Markovian dynamics. We also consider tree-like hierarchies of parent chains with parents coupling multiple hidden lower-level parent chains. We call this graphical model a dynamical system tree (DST) (see Figure 3(a)).

For example, multiple agents can interact and be aggregated by writing messages on some form of common bulletin board which evolves with hidden Markovian dynamics. Alternatively, a mediator or aggregator state could represent a coach directing a team of players or a script guiding multiple actors. One advantage of such a topology is that it has few parameters which reduces the chance of over-fitting. Conversely, a full HMM (as well as coupled and factorial HMMs to a certain extent) over all interacting agents must model the cross-product of their individual states which leads to an exponential number of parameters. Furthermore, the DST's mediated interaction approach lends itself nicely to hierarchical extensions. DSTs can have aggregators that themselves aggregate lower-level medi-ating chains to allow multiple scales of influence or layers of interaction. For instance, when modeling a university one may describe interacting people by mediating chains representing their research groups which are in turn aggregated and mediated by various departments, then schools and then a single mediating hidden state representing the evolution of the university as a whole. Alternatively, one may model the dynamics of a human, via a hierarchy over the individual limbs, fusing into upper and lower torso, etc. Admittedly, mediating variables could be reinterpreted as children (not parents) of the individual dynamical systems, but we prefer the DST's *mediating-parent* style of establishing interaction between processes. It avoids moralizing large cliques during inference, is nicely compatible with structured mean field derivations, and permits estimation of model parameters for an arbitrary tree hierarchy of interaction.

To construct a DST's probability distribution, we start from the bottom up by first considering a collection of simple independent dynamical systems we call leaf-processes. These individual systems are either continuous linear dynamical systems, or discrete HMMs or a hybrid as in a switching linear dynamical system (SLDS). Without loss of generality, we will assume all leaf-processes are SLDSs (as in [9] and in Figure 2) since these basically subsume both HMMs and Kalman filters. Furthermore, we assume that transitions between continuous hidden states are given by conditioned Gaussians and that emissions are continuous vectors from a Gaussian distribution given the continuous hidden state. On their own, the individual SLDSs do not capture the interactive nature of group dynamical behavior. To couple individual leaf-processes and model complex interaction, we have a hierarchy of aggregating Markovian processes that couple leaf-processes (or lower level aggregating-processes) as their children. Each aggregating process $a$ is denoted by its discrete Markovian hidden variables $s^a$ and defined as follows:

**Definition 1** *An aggregating-process is a Markov chain of hidden discrete states with at most one parent process and one or more children processes.*

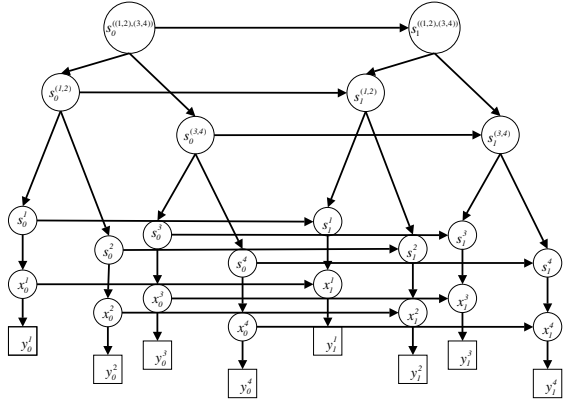

Figure 3: Dynamical Systems Tree (unrolled in time).

Children processes may be either other aggregating-processes themselves or leaf-processes. An aggregating-process' states are denoted by $s^a = \{s_0^a, \ldots, s_T^a\}$. Given its (possibly null) parent process $\pi(a)$ which has discrete hidden states $s^{\pi(a)} = \{s_0^{\pi(a)}, \ldots, s_T^{\pi(a)}\}$ the aggregating-process has the following conditional distribution:

$$p\left(s^a | s^{\pi(a)}\right) = p\left(s_0^a | s_0^{\pi(a)}\right) \prod_{t=1}^{T} p\left(s_t^a | s_{t-1}^a, s_t^{\pi(a)}\right).$$

The hierarchy of aggregating-processes is terminated by leaf-processes which contain both discrete and continuous hidden Markov dynamics as well as the actual emission or observation variables which we specify as follows:

**Definition 2** *A leaf-process is a switching linear dynamical system at the lowest level in the dynamical systems tree hierarchy. A leaf-process has at most one parent process and no children processes. The $i$'th leaf-process has discrete Markovian hidden states $s^i = \{s_0^i, \ldots, s_T^i\}$ as parents of continuous Markovian hidden states $x^i = \{x_0^i, \ldots, x_T^i\}$ as parents of independent emissions $y^i = \{y_0^i, \ldots, y_T^i\}$. Given its parent process $\pi(i)$ with discrete hidden states $s^{\pi(i)} = \{s_0^{\pi(i)}, \ldots, s_T^{\pi(i)}\}$ the leaf-process has the following conditional distribution:*

$$p(s^i, x^i, y^i | s^{\pi(i)}) = p(s_0^i | s_0^{\pi(i)}) p(x_0^i | s_0^i) p(y_0^i | x_0^i)$$
$$\times \prod_{t=1}^{T} p(s_t^i | s_{t-1}^i, s_t^{\pi(i)}) p(x_t^i | x_{t-1}^i, s_t^i) p(y_t^i | x_t^i).$$

Given $\mathcal{A}$ aggregating processes and $\mathcal{L}$ leaf-processes, the joint distribution $\mathcal{P}(\mathcal{S}, \mathcal{X}, \mathcal{Y})$ over all variables in the DST (namely $\{\mathcal{S}, \mathcal{X}, \mathcal{Y}\}$ which correspond to discrete hidden, continuous hidden and emission variables, respectively) is given by:

$$\mathcal{P}(\mathcal{S}, \mathcal{X}, \mathcal{Y}) = \prod_{a=1}^{\mathcal{A}} p(s^a | s^{\pi(a)}) \prod_{i=1}^{\mathcal{L}} p(s^i, x^i, y^i | s^{\pi(i)}).$$

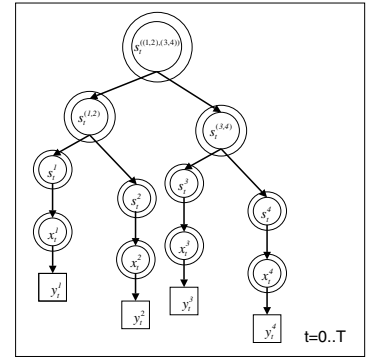

Figure 4: Dynamical Systems Tree (with replicators).

An example of a DST graphical model is shown in Figure 3 (unrolled time steps $t = 0 \ldots 1$). This DST has 4 leaf-processes and 3 aggregating-processes. The bottom aggregating processes $s^{(1,2)}$ and $s^{(3,4)}$ are parents of the pair of leaf processes in their superscripts. The bottom aggregating processes are themselves aggregated through one final parent process called $s^{((1,2),(3,4))}$. To avoid drawing DSTs unrolled in time, we represent them more compactly by only showing a single time instance of the DST at time $t$ and drawing a *replicator box* that indicates the network is repeated $t = 1 \ldots T$ times. Traditionally, replicator boxes show independent (iid) nodes (possibly linked to parent parameter nodes which we omit for clarity). We indicate Markovian dynamics between nodes in box $t-1$ to nodes in box $t$ by drawing extra replicator circles around all the nodes who inherit a link from their instantiation at the previous time step $t-1$. Nodes without the extra replicator circle (such as emission nodes) only have parents in the current replicator box $t$. Figure 4 depicts the DST in this compact replicator notation.

We now specify parameters for the aforementioned DST conditional distributions. The discrete distributions are multinomials while the continuous distributions are conditioned Gaussians. The parameters for the DST are:

$$p(s_0^a = j | s_0^{\pi(a)} = k) = \phi^a(j, k)$$
$$p(s_t^a = j | s_{t-1}^a = k, s_t^{\pi(a)} = l) = \Phi^a(j, k, l)$$
$$p(s_0^i = j | s_0^{\pi(i)} = k) = \psi^i(j, k)$$

$$p(s_t^i = j | s_{t-1}^i = k, s_t^{\pi(i)} = l) = \Psi^i(j,k,l)$$
$$p(x_0^i | s_0^i = j) = \mathcal{N}(x_0^i | \mu_j^i, q_j^i)$$
$$p(x_t^i | x_{t-1}^i, s_t^i = j) = \mathcal{N}(x_t^i | A_j^i x_{t-1}^i, Q_j^i)$$
$$p(y_0^i | x_0^i) = \mathcal{N}(y_0^i | C x_0^i, R)$$
$$p(y_t^i | x_t^i) = \mathcal{N}(y_t^i | C x_t^i, R).$$

Basic operations needed for DSTs include computing the likelihood of observations, inferring hidden states from an observation and estimating parameters from data. Essentially, EM learning and computing the likelihood hinge on performing inference over the hidden states. It is immediately evident that DST inference involves an intractable network since even the subcomponent SLDSs are intractable. Therefore we appeal to structured mean field for inference and perform approximate E-steps.

## 3  A STRUCTURED MEAN FIELD ALGORITHM

To avoid the intractabilities in the DST, we perform inference with a surrogate variational distribution that approximates our posterior $\mathcal{P}(\mathcal{S}, \mathcal{X} | \mathcal{Y})$ over the hidden variables given the observed data. We denote the simpler optimized surrogate distribution $\mathcal{Q}(\mathcal{S}, \mathcal{X})$ and display its conditional independence graph in Figure 5(a) unrolled in time for 3 time steps or in Figure 5(b) using replicator notation. This distribution resembles $\mathcal{P}$ except that all Markov chains are unlinked from each other which allows for efficient computation of marginal distributions and expected sufficient statistics.

Given a current setting of all our model parameters $\Theta$ and observation sequences, $\mathcal{Y}$, we can update a variational distribution on our DST by using the elegant formalisms outlined by [6, 1, 4]. More specifically, we have the following inequality on the incomplete log-likelihood:

$$\log \mathcal{P}(\mathcal{Y}|\Theta) \geq \sum_{\mathcal{S}} \int_{\mathcal{X}} \mathcal{Q}(\mathcal{S}, \mathcal{X}) \log \frac{\mathcal{P}(\mathcal{S}, \mathcal{X}, \mathcal{Y}|\Theta)}{\mathcal{Q}(\mathcal{S}, \mathcal{X})} d\mathcal{X}$$

where the right hand side is denoted by $\mathcal{B}(\mathcal{Q}, \Theta)$ for short and is a variational bound that makes contact with the left hand side when $\mathcal{Q}(\mathcal{S}, \mathcal{X}) = \mathcal{P}(\mathcal{S}, \mathcal{X}|\mathcal{Y}, \Theta)$. Since $\mathcal{Q}$ is a simpler and more factorized distribution than $\mathcal{P}(\mathcal{S}, \mathcal{X}|\mathcal{Y}, \Theta)$, the bound will be lowered and in general can no longer make tangential contact. We instead find the optimal distribution, $\mathcal{Q}$, that is as close as possible to the posterior distribution in terms of Kullback-Leibler divergence $KL(\mathcal{Q}(\mathcal{S}, \mathcal{X}) \| \mathcal{P}(\mathcal{S}, \mathcal{X}|\mathcal{Y}))$. Taking the functional derivative of $\mathcal{B}(\mathcal{Q}, \Theta)$ and setting it equal to zero, we arrive at the general structured mean field update equation for an independent chain[4]:

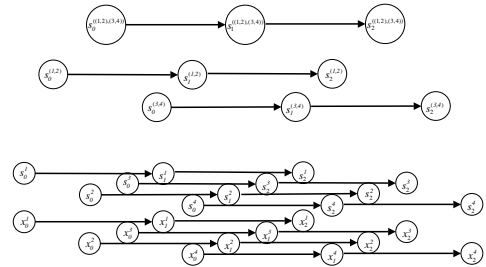

(a) Unrolled 3 Time Steps

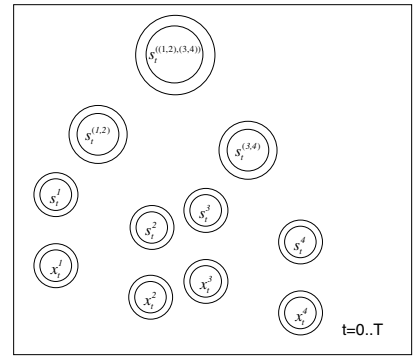

(b) Using Replicators

Figure 5: A Variational Distribution for DSTs.

$$\mathcal{Q}_i(\mathcal{S}_i) = \frac{1}{Z_i} e^{E_Q \{\log \mathcal{P}(\mathcal{S}, \mathcal{X}, \mathcal{Y}) | \mathcal{S}_i\}}$$

where $E_Q\{\cdot | \mathcal{S}_i\}$ is the conditional expectation with respect to all variables in $\mathcal{Q}$ except $\mathcal{S}_i$, $Z_i$ is a normalizing factor and substituting $a$ for $i$ and $\mathcal{X}$ for $\mathcal{S}$ where appropriate when updating each chain.

If we explicitly write the variational parameters of $\mathcal{Q}$ as:

$$\mathcal{Q}(s_0^a = j) = \hat{\phi}^a(j)$$
$$\mathcal{Q}(s_t^a = j | s_{t-1}^a = k) = \hat{\Phi}_t^a(j,k)$$
$$\mathcal{Q}(s_0^i = j) = \hat{\psi}^i(j)$$
$$\mathcal{Q}(s_t^i = j | s_{t-1}^i = k) = \hat{\Psi}_t^i(j,k)$$
$$\mathcal{Q}(x_0^i) = \mathcal{N}(x_0^i | \hat{\mu}^i, \hat{q}^i)$$
$$\mathcal{Q}(x_t^i | x_{t-1}^i) = \mathcal{N}(x_t^i | \hat{A}_t^i x_{t-1}^i + \hat{B}_t^i, \hat{Q}_t^i)$$

we get the following update rules:

$$\hat{\Phi}_t^a(j,k) \propto \exp\Big(\sum_l \langle s_t^{\pi(a)}(l) \rangle \log \Phi^a(j,k,l)$$
$$+ \sum_{c \in \text{Child}(a)} \sum_{h,i} \langle s_t^c(h) s_{t-1}^c(i) \rangle \log \Phi^c(h,i,j) \Big)$$

$$\hat{\phi}^a(j) \propto \exp\Big(\sum_l \langle s_t^{\pi(a)}(l) \rangle \log \phi^a(j,l)$$
$$+ \sum_{c \in \text{Child}(a)} \sum_h \langle s_0^c(h) \rangle \log \Phi^c(h,j) \Big)$$

$$\hat{\Psi}_t^i(j,k) \propto \exp\Big(\sum_l \langle s_t^{\pi(i)}(l) \rangle \log \Psi^i(j,k,l) - \tfrac{1}{2} \log |Q_j^i|$$

$$\begin{aligned}
&\quad -\tfrac{1}{2}\langle (x_t^i - A_j^i x_{t-1}^i)'(Q_j^i)^{-1}(x_t^i - A_j^i x_{t-1}^i)\rangle\big) \\
\hat{\psi}^i(j) &\propto \exp\Big(\sum_l \langle s_t^{\pi(i)}(l)\rangle \log \psi^i(j,l) - \tfrac{1}{2}\log|q_j^i| \\
&\quad -\tfrac{1}{2}\langle (x_0^i - \mu_j^i)'(q_j^i)^{-1}(x_0^i - \mu_j^i)\rangle\Big) \\
\hat{A}_t^i &= \hat{Q}_t^i \sum_j \langle s_t^i(j)\rangle (Q_j^i)^{-1} A_j^i \\
\hat{B}_t^i &= \hat{Q}_t^i (C'R^{-1}y_t^i + (\hat{A}_{t+1}^i)'(\hat{Q}_{t+1}^i)^{-1}\hat{B}_{t+1}^i) \\
(\hat{Q}_t^i)^{-1} &= \sum_j \langle s_t^i(j)\rangle (Q_j^i)^{-1} + \sum_j \langle s_{t+1}^i(j)\rangle (A_j^i)'(Q_j^i)^{-1}A_j^i \\
&\quad -(\hat{A}_{t+1}^i)'(\hat{Q}_{t+1}^i)^{-1}\hat{A}_{t+1}^i + C'R^{-1}C \\
\hat{\mu}^i &= \hat{q}^i \sum_j \langle s_0^i(j)\rangle (q_j^i)^{-1}\mu_j^i \\
&\quad +\hat{q}^i (C'R^{-1}y_1^i + (\hat{A}_2^i)'(\hat{Q}_2^i)^{-1}\hat{B}_2^i) \\
\hat{q}^i &= \sum_j \langle s_0^i(j)\rangle (q_j^i)^{-1} + \sum_j \langle s_1^i(j)\rangle (A_j^i)'(Q_j^i)^{-1}A_j^i \\
&\quad -(\hat{A}_1^i)'(\hat{Q}_1^i)^{-1}\hat{A}_1^i + C'R^{-1}C
\end{aligned}$$

where $\langle \cdot \rangle$ denotes the expectation with respect to $\mathcal{Q}$.

We use these variational parameters to compute the marginal distributions and expectations over the hidden variables. Because $\mathcal{Q}$ is a set of disconnected chains, we need only the following expected sufficient statistics: $\langle x_t\rangle$, $\langle x_t, x_t\rangle$, $\langle x_t, x_{t-1}\rangle$, $\langle s_t\rangle$ and $\langle s_t, s_{t-1}\rangle$ for parameter estimation in the M-step. The terms for the discrete chains, $\langle s_t\rangle$ and $\langle s_t, s_{t-1}\rangle$, can be computed efficiently from the unnormalized $\hat{\Phi}_t^a(j,k)$, $\hat{\phi}^a(j)$, $\hat{\Psi}_t^i(j,k)$ and $\hat{\psi}^i(j)$ with a simple forward-backward inference algorithm. The terms for the continuous chains, $\langle x_t\rangle$, $\langle x_t, x_t\rangle$, $\langle x_t, x_{t-1}\rangle$ can be computed directly from the fully specified distributions $\mathcal{N}(x_t^i|\hat{A}_t^i x_{t-1}^i + \hat{B}_t^i, \hat{Q}_t^i)$ and $\mathcal{N}(x_0^i|\hat{\mu}^i, \hat{q}^i)$ using a recursion based on identities for conditioned Gaussians:

$$\begin{aligned}
\langle x_1^i\rangle &= \hat{\mu}^i \\
\langle x_1^i, x_1^i\rangle &= \hat{q}^i + \langle x_1^i\rangle \langle (x_1^i)'\rangle \\
\langle x_t^i\rangle &= \hat{A}_t^i \langle x_{t-1}^i\rangle + \hat{B}_t^i \\
\langle x_t^i, x_t^i\rangle &= \hat{A}_t^i \langle x_{t-1}^i, x_{t-1}^i\rangle (\hat{A}_t^i)' + \hat{Q}_t^i \\
\langle x_t^i, x_{t-1}^i\rangle &= \hat{A}_t^i \langle x_{t-1}^i, x_{t-1}^i\rangle
\end{aligned}$$

where, again, $\langle \cdot \rangle$ denotes the expectation with respect to $\mathcal{Q}$. If covariance matrices are needed, they can easily be computed from the sufficient statistics.

Previous algorithms for performing variational inference in related hybrid dynamic Bayes nets such as the switching state space model [2] and the switching linear dynamical system [9] propose forming a Kalman filter for the hidden Gaussian chains and using RTS smoothing to perform inference. This requires both a forward and backward recursion after the initial structured mean field recursion for a total of three recursions. Our method, which is also applicable to these models, only requires two recursions: a backward recursion to form the variational distribution and a forward recursion to compute the expected sufficient statistics. This is more efficient and more numerically stable. Numerical stability becomes important as the number of time steps increases. Kalman filters (and similar covariance based recursions) can suffer from numerical errors where covariances can become no longer symmetric positive definite. So, by reducing the number of covariance recursions, we reduce the propagation of numerical errors.

```
init variational parameters
forward-backward inference on all chains
compute  B(Q⁰,Θ)
while  (B(Q^{ι+1},Θ) − B(Q^ι,Θ) ≥ ∈)
{   recurse through tree hierarchy
      for each chain in recursion
         iterate updates of variationals
         do forward-backward inference
   update  B(Q^{ι+1},Θ)   for all chains     }
```

Figure 6: Pseudo Code for Updating Variational $\mathcal{Q}$.

After the above approximate E-step, an M-step update of the parameters $\Theta$ is trivial via expectations of the complete likelihood using the current $\mathcal{Q}$ distribution. The update rules for the model parameters for a Kalman filter (with continuous dynamics and continuous emission models) are shown in [1]. Similarly, updating the discrete Markov chain's transition table is immediate.

Computing the model's true log-likelihood, however, remains intractable. We instead evaluate the bound, $\mathcal{B}(\mathcal{Q},\Theta)$. During learning, the bound increases monotonically as we iterate variational inferences in $\mathcal{Q}$ and model parameter updates in $\Theta$ to achieve a local maximum. We compute the bound using expectations with respect to $\mathcal{Q}$: $\mathcal{B}(\mathcal{Q},\Theta) = E_{\mathcal{Q}(\mathcal{S},\mathcal{X})}\{\log \mathcal{P}(\mathcal{S},\mathcal{X},\mathcal{Y}) - \log \mathcal{Q}(\mathcal{S},\mathcal{X})\}$. These expectations are easy to compute and only involve expectations over (at most pairwise) cliques of $\mathcal{Q}$. In all the above computations, it is easy to recurse through the DST to compute the $\mathcal{B}(\mathcal{Q},\Theta)$ terms for each leaf-process and aggregator-process. Furthermore, variational parameter updates are nicely decoupled and M-steps for $\Theta$ parameters are independent given the inferred expectations under the $\mathcal{Q}$ distribution. In Figure 6 we show pseudo-code for recursing through the hierarchy tree to update variational parameters. This is interleaved with re-estimation of the model parameters.

## 4   EXPERIMENTS

To demonstrate their diverse applicability, we evaluated DSTs and compared them with other dynamical models on gene expression levels for cancerous hu-man cells [12], and on real-world trajectory data from American football plays [3]. Models were trained using our variational EM algorithm. We used the over-relaxed bound optimization technique [10] to speed up convergence. In practice this required three to four times fewer EM steps to converge than our standard algorithm implemented without this technique.

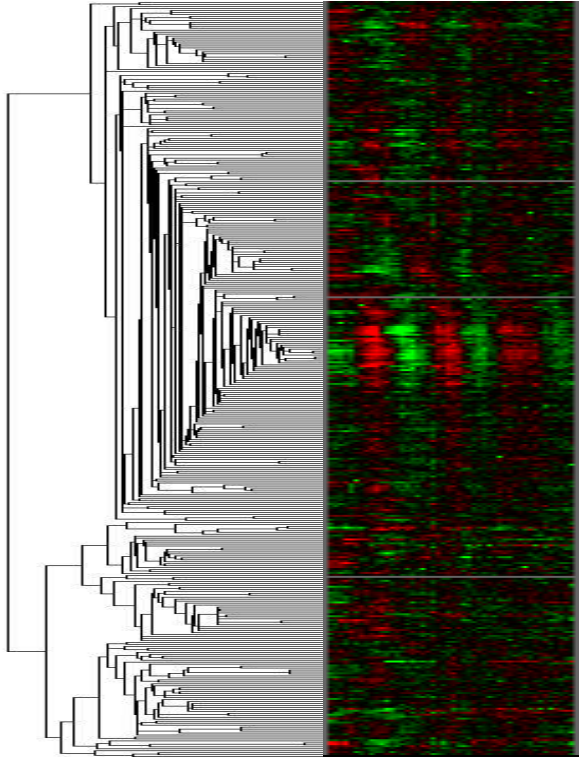

Figure 7: Gene expression data and clustering.

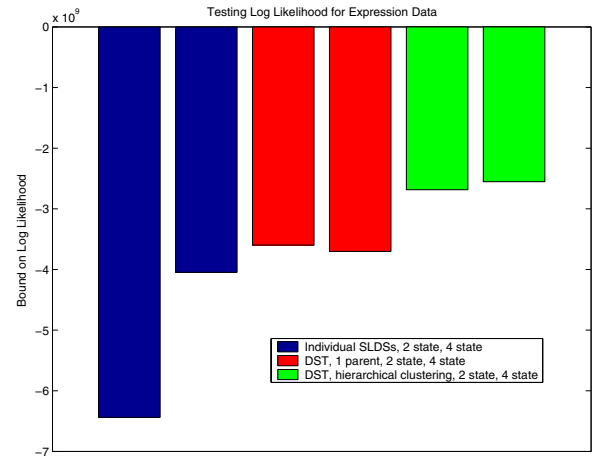

Figure 8: Bound on Log Likelihood for Previously Unseen Gene Testing Data

### 4.1 GENE EXPRESSION LEVELS

Genes active in cancerous cells are identified in [12]. We look at a subset of 384 that were studied in the paper. Each gene expression level is a 1 dimensional time series, depicted in Figure 7, representing the gene's deviation from the baseline. In these experiments we wish to compare genes modeled independently to genes modeled as a group with various DST topologies. We train our models on the time series identified as Thy-Thy 3 (named for the method of synchronization used). This series is the most thoroughly measured data set available from these sets of experiments. There are 45 time steps separated by 1 hour which span 3 cell cycles. We then test our learned models on the first 2 cell cycles of the Thy-Thy 2 data set that was obtained under similar experimental conditions. Model goodness is evaluated based on testing log likelihood (evaluated as a bound). This testing data is more sparsely sampled with gaps of 2 hours toward the end. In order to maintain synchronicity with the models, we infer the missing data along with the other hidden variables.

We first considered modeling the genes with 384 individual SLDSs. The hidden continuous state dimensionality was chosen to be 1 and both 2 and 4 discrete switching states were considered. The second model considered was a naive DST with one parent aggregating process and 384 child leaf SLDSs. Again, the continuous state dimensionality was chosen to be 1 and the discrete states for both the aggregating process and leaf processes were evaluated for both 2 and 4 states. The final model considered is a complicated DST based on the hierarchical clustering reported in [12] and depicted in Figure 7. The DST has 383 aggregating processes and 384 leaf SLDSs. The parameter dimensionality was chosen to be the same as the previous models. Because parameter learning using EM is plagued with local maxima, good initialization is quite important. To initialize the continuous variables, we trained a linear dynamic system for each state in the SLDS leaf process using contiguous non-overlapping subsets of the data. All discrete variables were initialized to a slight random perturbation away from the maximum entropy distribution. In practice, this yielded good results and allowed for symmetry breaking without overly biasing the switching states a priori. Results are reported in Figure 8. From left to right, the graph shows SLDSs, naive DST and clustered DST each with 2 and 4 switching states. Note that larger bars are worse (more negative). DSTs demonstrate superior ability to fit data in this domain. More importantly, the complicated DST, although having many more parameters to fit, does not over-fit and performs best. This intuitively suggests that good knowledge of the domain encoded in the structure of the DST will improve the model.

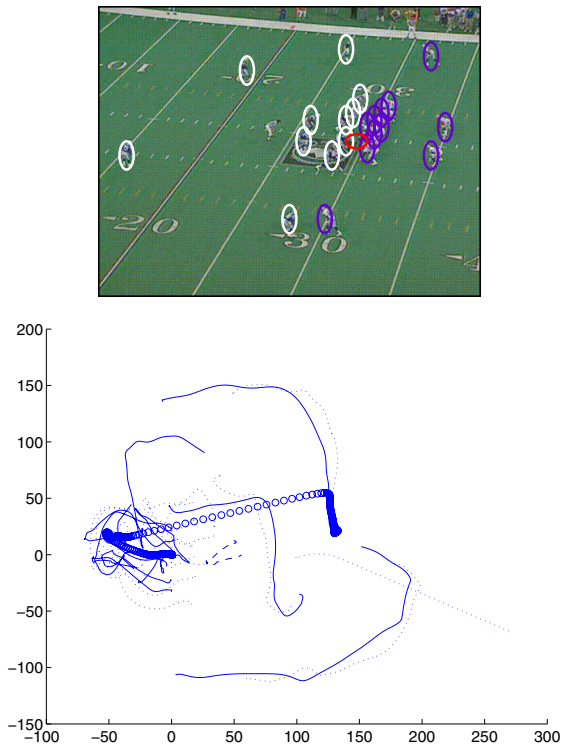

Figure 9: Player Trajectory Data Gathered from Video.

### 4.2 FOOTBALL TRAJECTORIES

In [3], players are tracked using computer vision to obtain spatial coordinates on the football field (with some normalization). Each human generates a continuous time series of two dimensional coordinates. We wish to compare DSTs for model accuracy and classification. Our training data consists of 7 examples of a passing play called *dig* and 7 examples of a running play called *wham* recorded from a real football game. Testing was performed using leave one out cross validation. Figure 9 shows the trajectories for multiple players during a play. In order to ensure that classification is based on player's dynamics and not on their starting position, the data was translated so that each players' trajectory is their offset from the origin.

A naive approach to modeling our data is to stack or concatenate each player's time series into a single multivariate series. We start with a simple SLDS modeling all of the players with 4 dimensions for the continuous hidden states and 4 discrete switching states. To instead model players as separate temporal processes, we then trained multiple independent SLDSs for each player in isolation. Each has 2 dimensional continuous $x_t$ state and 2 states for the switches $s_t$. Yet such SLDSs completely ignore interactions between players. A DST, however, can couple many separate temporal interactions by fusing SLDS structures. We consider two topologies. The first consists of a single binary-state game-chain aggregating all of the players in one group. The second consists of two additional binary-state team-chains aggregating the two teams of players and a top level binary-state game-chain aggregating the two teams. Parameters were initialized in the same manner that is described in the previous experiment. Results for classification and average (bound on) testing log likelihood are reported in the table below. Testing results show that treating the players as individual time series, as opposed to a concatenation, yields better models based on likelihood. The more complicated DST slightly outperforms the individual SLDS in classification and on the average (bound on) testing likelihood for wham plays. However the less complicated DST performed far worse than the more complicated one. This again provides evidence to our intuition that DSTs built on sound domain knowledge will perform well. Better results for a DST model could possibly be obtained if more domain knowledge is encoded in the structure. For example, players may be grouped based on their role on the field, such as offensive line and defensive line, before being grouped as a team.

| Model | LL Test Dig | LL Test Wham | Errors |
|---|---|---|---|
| Single SLDS | -3.1143e+5 | -1.1826e+5 | 4 |
| Multi SLDSs | -6.4109e+3 | -5.3838e+3 | 1 |
| DST1 | -8.4900e+3 | -5.5048e+3 | 5 |
| DST2 | -6.6841e+3 | -5.1342e+3 | 0 |

## 5 Discussion

We presented dynamical systems trees as an alternative dynamical Bayesian network model specifically for coupling multiple interacting processes (i.e. individual switched linear dynamical systems) via a tree structured hierarchy of influence. The hidden states of individual dynamical systems are modeled as children of an aggregating process and further aggregating process parents continue the grouping at increasingly looser and higher levels of interaction. Restricting mediator or aggregator chains to be parents of lower-level interacting chains helps maintain a compact model with a small number of parameters. Furthermore, due to the tree hierarchy of the aggregating processes, structured mean field computations become formulaic and recursive and can be easily implemented for an arbitrary topology of the DST. Experiments demonstrated their ability to model two difficult real world problems. The importance of properly encoding domain knowledge in the DST topology was demonstrated. This is to be expected of any Bayes net, but the ease of adjusting the topology without needing to re-derive or re-code new algorithms makes it simple to explore different DST structures. Future work will attempt to automatically group time series to determine the best DST topology. Simple clustering had good results with the gene expression data, but a more well motivated method based on likelihood is desirable. Overall, dynamical systems trees provide an easily reconfigurable and intuitive architecture for modeling temporal interaction data.

## Acknowledgements

We would like to thank Anshul Kundaje and Darrin Lewis for helpful discussions on gene expression levels. This research is funded in part by grant IIS-0347499 from the NSF.